\documentclass[12pt, letterpaper]{article}
\usepackage{graphicx}
\usepackage{cite}
\usepackage{picinpar}
\usepackage{amsmath}
\usepackage{url}
\usepackage{flushend}
\usepackage[latin1]{inputenc}
\usepackage{colortbl}
\usepackage{soul}
\usepackage{multirow}
\usepackage{pifont}
\usepackage{color}
\usepackage{alltt}
\usepackage[hidelinks]{hyperref}
\usepackage{enumerate}
\usepackage{siunitx}
\usepackage{breakurl}
\usepackage{epstopdf}
\usepackage{pbox}

\usepackage{tabularx,booktabs}

\usepackage{amssymb}

\title{Online Recognition of Incomplete Gesture Data to Interface Collaborative	Robots
	
	\thanks{\textit{\underline{Citation}}: 
		\textbf{M. A. Simão, O. Gibaru and P. Neto, "Online Recognition of Incomplete Gesture Data to Interface Collaborative Robots," in IEEE Transactions on Industrial Electronics, vol. 66, no. 12, pp. 9372-9382, Dec. 2019, doi: 10.1109/TIE.2019.2891449.}} 
}

\author{
	Miguel Simão, Olivier Gibaru, P. Neto \\
	University of Coimbra, Portugal \\
}

\begin{document}
	\maketitle

\thanks{
	
	{\color{black}
		Miguel Simão and Pedro Neto are with the Department of Mechanical Engineering, University of Coimbra, 3030-788 Coimbra, Portugal (e-mail: miguel.simao@uc.pt; pedro.neto@dem.uc.pt). 
		
		Olivier Gibaru is with the Ecole Nationale Supérieure d'Arts et Métiers, ParisTech, 59800 Lille, France (e-mail: olivier.gibaru@ensam.eu).
	}
}
	
\begin{abstract}
Online recognition of gestures is critical for intuitive human-robot
interaction (HRI) and further push collaborative robotics into the
market, making robots accessible to more people. The problem is that
it is difficult to achieve accurate gesture recognition in real unstructured
environments, often using distorted and incomplete multi-sensory data.
This paper introduces a HRI framework to classify large vocabularies
of interwoven static gestures (SGs) and dynamic gestures (DGs) captured
with wearable sensors. DG features are obtained by applying data dimensionality
reduction (DDR) to raw data from sensors (resampling with cubic interpolation
and principal component analysis (PCA)). Experimental tests were conducted
using the UC2017 hand gesture dataset with samples from eight different
subjects. The classification models show an accuracy of 95.6\% for
a library of 24 SGs with a random forest (RF) and 99.3\% for 10 DGs
using artificial neural networks (ANNs). These results compare equally
or favourably with different commonly used classifiers. Long Short-Term
Memory (LSTM) deep networks achieved similar performance in online
frame-by-frame classification using raw incomplete data, performing
better in terms of accuracy than static models with specially crafted
features, but worse in training and inference time. The recognized
gestures are used to teleoperate a robot in a collaborative process
that consists in preparing a breakfast meal.
\end{abstract}


\section{Introduction}

The paradigm for robot usage has changed in the
last few years, from an idea in which robots work with complete autonomy
to a scenario in where robots cognitively collaborate with human beings.
This brings together the best of each partner, robot and human, by
combining the coordination and cognitive capabilities of humans with
the robots' accuracy and ability to perform monotonous tasks. Robots
and humans have to understand each other and interact in a natural
way (using gestures, speech and physical interaction), creating a
co-working partnership. This will allow a greater presence of robots
in all domains of our society. The problem is that the existing interaction
modalities are neither intuitive nor reliable. Instructing and programming
an industrial robot by the traditional teaching method is a tedious
and time-consuming task that requires technical expertise in robot
programming. 

The collaborative robotics market is rapidly growing and HRI interfaces
have a main role in the acceptance of robots as partners. Gestures
are an intuitive interface to teleoperate a robot since they intuitive
to use and do not require technical skills in robot programming to
be used \cite{doi:10.1177/0278364917709941}. For instance, a human
co-worker can use a DG to indicate a grasping position and use a SG
to stop the robot \cite{7793333}. In this scenario, the human has
little or nothing to learn about the interface, focusing instead on
the task being performed. The robot assists the human when necessary,
thus reducing the exposition to poor ergonomic conditions and injury.

\begin{figure}[!t]\centering
	\includegraphics[width=11.5cm]{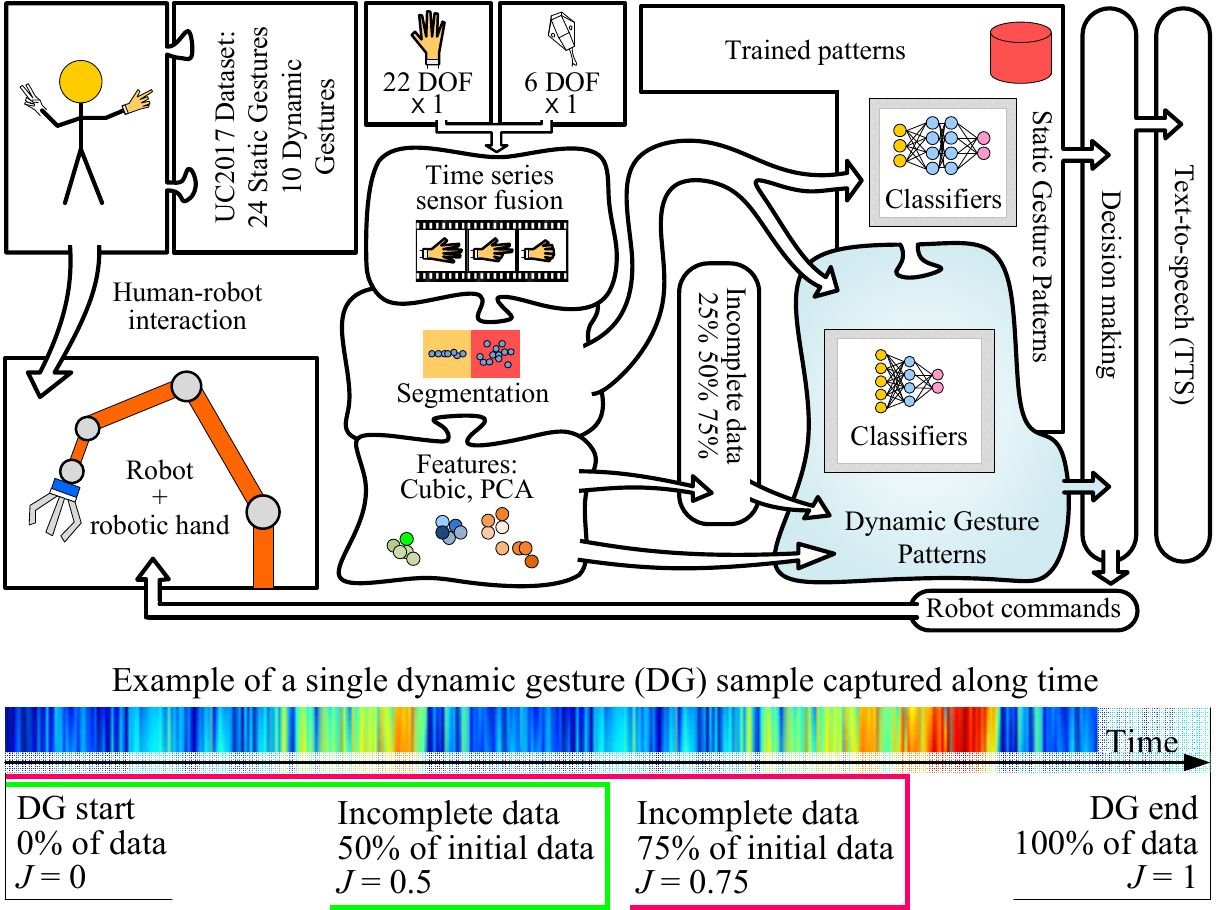}
	
	\caption{Overview of the proposed gesture-based HRI framework: data acquisition,
		segmentation, features, classification and the robot interface. At
		the bottom it is explained the meaning of incomplete data for DG classification.
		For example a DG can be classified with initial 50\% ($J=0.5$) of
		data representing such gesture, i.e., DGs can be classified in anticipation,
		before the user finishes the gesture in real world.} \label{fig:INTRO-Segmentation-1}
\end{figure}

This paper proposes an integrated modular gesture-based HRI framework,
Fig. \ref{fig:INTRO-Segmentation-1}. Static and dynamic gesture segments,
composed of data captured by a data glove and magnetic tracker, are
created automatically with a motion detection algorithm applied to
a sliding window. Static segments are used as input for SG classifiers.
Dynamic segments, which are discriminated by DG classifiers, are subject
to data dimensionality reduction (DDR) with resampling based on cubic
interpolation (CI) or principal component analysis (PCA). Traditional
probabilistic latent variable models, such as PCA, are static linear
approaches in which the dynamics and nonlinearities are not properly
considered. In \cite{7997769}, the authors propose a weighted linear
dynamic system (WLDS) for nonlinear dynamic feature extraction which
showed superior prediction accuracy. Nevertheless, the real-time performance
is worse when compared to static approaches because it requires more
computational time. This is undesirable and limits the online performance
of the proposed gesture-based HRI system. The proposed CI and PCA
approaches are demonstrated to be computationally inexpensive without
sacrificing classification accuracy, even when used with incomplete
data. We use large vocabularies of gestures (a total of 34) and a
relatively low number of training samples to simplify and expedite
the training process. Experiments demonstrated that standard classifiers,
such as artificial neural networks (ANNs), are reliable in both SG
and DG classification. Furthermore, they compare equal/favorably to
deep learning classifiers (LSTM and Convolutional Neural Networks
(CNNs)) in both inference time and accuracy. Finally, a robot task
manager maps the classified gestures to robot commands.

\subsection{Motivation, Challenges and Contributions}

A major challenge is the continuous, online and reliable recognition
of gestures from real-time data streams. Continuous gesture recognition
is the natural way used by humans to communicate with each other,
in which communicative gestures (the effective SGs and DGs with an
explicit meaning) appear intermittently with non-communicative gestures
(pauses and movement epenthesis (ME) \textendash{} inter-gesture transition
periods) in a random order \cite{Yang2010}. Many studies do not approach
gesture classification in this continuous manner, nor address the
negative effect of ME. It is also a challenge to recognize gesture
patterns from incomplete data, as well as intuitively map the recognized
gestures into robot commands for natural and safe HRI. In this context,
the motivations behind this study are:
\begin{enumerate}[1)]
	\item Combine and fuse sensor data from multiple wearable devices in order
	to capture a person's gestures (hand and arms) accurately, without
	occlusions;
	\item Application of proper DDR methods to increase classification accuracy,
	reduce the training time, reduce the number of samples required to
	train the classifiers, while allowing online implementation;
	\item Achieve high recognition rates (close to 100\%) and ensure generalization
	capability in respect to untrained samples and new users;
	\item Classification of DGs from incomplete data, allowing the classification
	of a gesture while it is being performed by the human;
	\item Intuitive and online interfacing with a robot using gestures.
\end{enumerate}
The proposed system was evaluated by conducting several experiments
using wearable/body-worn sensors (a data glove and a magnetic tracker),
resulting in the following contributions:
\begin{enumerate}
	\item The combination of DDR (CI and PCA) and ANNs for DG classification
	from wearable sensor data resulted in high classification accuracy
	that compares favorably with standard classifiers, including deep
	learning LSTM and CNNs. This method is computationally inexpensive,
	allowing the gesture-based online interaction with the robot. Gestures
	are recognized with an accuracy of 95.6\% for a library of 24 SGs
	and 99.3\% for 10 DGs. 
	\item The above results were obtained in continuous data, with multiple
	subjects (user independence) and applied in an unstructured environment;
	\item Sequential classification of DGs showed an accuracy that is higher
	with incomplete data (50\% or 75\% of initial frames of data that
	represent a DG) than with 100\% of DG data across different classifiers
	and users. In this context, DGs can be classified in anticipation,
	before the user finishes a gesture;
	\item Framework tested in real unstructured environment where the recognized
	gestures serve as an intuitive interface to manage an online collaborative
	process, in which a robot assists the human in the preparation of
	a breakfast meal.
\end{enumerate}

\subsection{Related Work}

Gesture-based HRI for collaborative robotics is an emerging
and multidisciplinary research field. Communicative gestures provide
information that is difficult to convey in speech, i.e., command gestures,
pointing, gestures addressed to objects or actions, and mimicking
gestures \cite{Burke2015,Waldherr2000}. Gestures have been proven
to be one of the most effective and natural mechanisms for reliable
HRI \cite{7835143}. They have been used for robot teleoperation
and to coordinate the interactive process of cooperation between human
and robot. As stated in \cite{Ende2011}, an interactive robotic task
generally consists of individual actions, operations and motions that
are arranged in a hierarchical order so that the process can be managed
by simple gestures \cite{indin2018}. 

An inefficient segmentation process (determining when a gesture starts
and ends) results in a classification model that is more likely to
fail \cite{Simao2016}. The analysis of continuous data streams to
solve spatial and temporal segmentation is challenging \cite{Alon2009}.
The problem is that it is difficult to automate the segmentation process,
making gesture recognition in real world scenarios a difficult task
\cite{SegmentationIECON2016}. 

The input features for gesture recognition are normally the hand/arm/body
position, orientation and motion \cite{Yang2010}, often captured
from vision sensors. Owing to its naturalness, in opposition to wearable
sensors that need to be attached to the human body, vision sensing
is the most common interaction technology. Gesture classification
from video stream requires large amounts of training data, especially
for state-of-the-art deep learning classifiers. Moreover, it is difficult
to construct reliable features from only vision sensing due to occlusions,
varying light conditions and free movement of the user in the scene
\cite{Burger2011,Waldherr2000}. To improve classification reliability,
a significant number of studies combine data from vision and wearable
sensors. Taking this into account, several approaches to gesture recognition
rely on wearable sensors such as data gloves, magnetic tracking sensors,
inertial measurement units (IMUs) and electromyography (EMGs), among
others. In fact, these interaction technologies have been proven to
provide reliable features in unstructured environments. Nevertheless,
they also place an added burden on the user since they are worn on
the body.

Some gestures, although not all, can be defined by their spatial trajectory,
e.g., a circle. Burke and Lasenby succeeded on using PCA and Bayesian
filtering for the classification of time series gestures \cite{Burke2015}.
Hidden Markov Models (HMMs) can be used to find time dependencies
in skeletal features extracted from image and depth data (RGB-D) with
a combination of Deep Belief Networks (DBNs) and 3D CNNs \cite{Wu2016}.
Deep learning combined with recurrent LSTM networks demonstrated state-of-the-art
performance in the classification of human activities from wearable
sensors \cite{Roggen2016DEEP}. Features are automatically extracted
from raw sensor data, avoiding the need for expert knowledge in feature
design. The reported results show that this framework outperforms
competing deep non-recurrent networks. Various ANNs in series demonstrated
superior performance in the classification of a high number of gesture
classes \cite{Neto2013}. Field et al. used a Gaussian Mixture Model
(GMM) to classify human body postures with previous unsupervised temporal
clustering \cite{Field2015}. Switching Gaussian Process Dynamic Models
(SGPDM) are proposed to capture motion dynamics and to identify motion
classes such as walk or run, and smile or angry \cite{5206580}. Recognition
performance on real videos (comparatively low quality, low frame rates
and with pose changes) demonstrated that the SGPDM model can efficiently
track composite motions with various dynamics. A framework for dynamic
hand gesture recognition using Generalized Time Warping (GTW) for
alignment of time series is proposed in \cite{10.1007/978-3-319-27863-6_64}.
Features are extracted from the aligned sequences of hand gestures
based on texture descriptors, and the hand motion recognition is performed
by CNNs. Autoencoders and stacked autoencoders (SAE) have been successfully
used for feature representation in various applications \cite{8302941}.

Recent studies report state-of-the-art methods for hand detection
and gesture classification from RGB-D video using deep learning \cite{Monnier2015}.
Generally speaking, deep learning requires large amounts of training
data, the models are computationally expensive to train, and it is
challenging to determine good hyperparameters, since deep networks
are essentially black boxes (it is difficult to know exactly how and
why they output certain values). Boosting methods, based on ensembles
of weak classifiers, allow multi-class hand detection \cite{Mei2015}.
Despite all the proposed solutions, it is still challenging to use
of gestures as a reliable interaction modality to control a robot/machine
in real-time.

The evolution of pattern recognition has been enormous in the last
few years. However, many of existing solutions address object or SG
classification which is less challenging than sequential classification.
Results obtained in well-established datasets have good accuracy in
offline classification but are seldom tested online and the processing
time is not mentioned. The ability to classify a gesture online is
critical to interface with a machine/robot. Finally, no studies approach
gesture classification from incomplete data, being normally assumed
that more data results in better accuracy.

\section{Gesture Classification}

\subsection{Problem Formulation}

Within a continuous data stream, there may be a sequence of SG
and DG with no specific order. As new frames are acquired,
they are segmented into static or dynamic frames with a motion detection
algorithm \cite{Simao2016}. This algorithm identifies motion, or
lack thereof, including sudden inversions of movement direction which
are common in DGs. This is achieved by the analysis of velocities
and accelerations numerically derived from positional data. A genetic
algorithm is used to compute motion thresholds from calibration data.
As a result, we have static and dynamic blocks of frames contiguous
in time. Static blocks are SG candidates and dynamic blocks are DG
candidates. Therefore, we propose two independent classifiers, one
for the classification of SGs and the other for the classification
of DGs.

The segmentation function $\varGamma$ based on a motion-threshold
algorithm is applied to a window of a stream of data $\boldsymbol{S}$,
of dimensionality $d$ and length $n$: $\left\{ \left(\boldsymbol{S},\varGamma(\boldsymbol{S})\right):\ \boldsymbol{S}\in\mathbb{R}^{d\times n}\ \text{and}\ \varGamma(\boldsymbol{S})\in\left\{ 0,1\right\} ^{n}\right\} $.
The static frames indicating no motion (input data for the SG classifier)
are defined by $m_{i}=0$ and the dynamic frames indicating motion
(input data for the DG classifier) by $m_{i}=1$.

\global\long\def\S{\mathcal{S}}
\global\long\def\R{\mathbb{R}}

The dynamic segments are extracted by a search function that finds
transitions in $m$ (from $0$ to $1$ and $1$ to $0$). Given two
consecutive transitions in the frames $i$ and $i+k$ so that $m_{i-1}=0$,
$m_{i}=1$, $m_{i+k-1}=1$ and $m_{i+k}=0$, a DG sample is
defined by:

\begin{equation}
	\mathbf{X}^{D}=\left[S_{\bullet i}\ S_{\bullet i+1}\ ...\ S_{\bullet i+k-1}\right],\quad\mathbf{X}^{D}\in\R^{d\times k}\label{eq:def-sample-dynamic}
\end{equation}
where the $S_{\bullet i}$ vector is the \emph{i}-th column (frame)
of the data stream. In terms of matrix notation, being $\mathbf{A}\in\mathbb{M}^{p\times q}$,
$\mathbf{A}_{ij}$ represents the element of the array $\mathbf{A}$
with row $i$ and column $j$, $\mathbf{A}_{i\bullet}\equiv\left[\mathbf{A}_{i1}\cdots\mathbf{A}_{in}\right]$
and $\mathbf{A}_{\bullet j}\equiv\left[\mathbf{A}_{1j}\cdots\mathbf{A}_{nj}\right]^{T}$,
and $\mathbb{M}$ is the notation for a real-valued matrix.

The static gesture samples are considered the first frame after a
transition from $m_{i+k-1}=1$ to $m_{i+k}=0$:

\begin{equation}
	\mathbf{X}^{S}=S_{\bullet i+k},\quad\mathbf{X}^{S}\in\R^{d}\label{eq:def-sample-static}
\end{equation}

Hereinafter, the notation for a sample independently of its nature
(static or dynamic) is $\mathbf{X}$. Static and dynamic samples are
differentiated by their dimensionality. The \emph{i}-th sample of
a dataset is represented by $\mathbf{X}^{(i)}$. We represent the
feature extraction pipeline by $\varPi$, which is used to transform
the raw data into the predictors $\mathbf{z}$ that feed the classifiers:
$\left\{ \left(\boldsymbol{X},\mathbf{z}=\varPi\left(\boldsymbol{X}\right)\right):\ X\in\mathbb{R}^{d\times n}\ \text{and}\ \mathbf{z}\in\mathbb{R}^{b}\right\} $,
where $d$ is the number of channels of the sample and $n$ its length.
The target vectors are one-hot encoded class indexes. For any given
sample, the target class has the index $o$ and the target vector
$\mathbf{t}^{(o)}$ is defined by $\mathbf{t}_{j}^{(o)}=\delta_{oj},\ j=1,...,n_{classes}$.
Therefore $\mathbf{t}\in\left\{ 0,1\right\} ^{n_{classes}}$, $\delta$
is the Kronecker delta and $\mathbf{t}_{j}$ is the \emph{j}-th element
of $\mathbf{t}$.

For DG, the transformation $\varPi$ could yield a long vector,
which often makes training the classifier more difficult. Therefore,
we introduce DDR at the end of the pipeline $\varPi$, such as PCA
and CI. The feature vectors are fed into the respective classifiers.

In this study, our aim is to map the classified gestures into robot
actions, such as moving to a target or halting movement. However,
before issuing an action command, we must exclude poorly classified
patterns from the stream. For example, we can exclude classifications
by context and by applying a threshold to the classification score:

\begin{equation}
	\mathbf{o}_{i}=\begin{cases}
		\text{SG}_{t}, & \text{if }p(\mathbf{y}=\mathbf{t}|\mathbf{z})\geq\tau^{S}\wedge m_{i}=0\\
		\text{DG}_{t}, & \text{if }p(\mathbf{y}=\mathbf{t}|\mathbf{z})\geq\tau^{D}\wedge m_{i}=1\\
		0, & \text{otherwise}
	\end{cases}
\end{equation}
where $\mathbf{o}_{i}$ is the output gesture class number, $p(\mathbf{y}=\mathbf{t}|\mathbf{z})$
is the likelihood of the classifier's output $\mathbf{y}$ being in
the class $\mathbf{t}$ given the $\mathbf{z}$ input, $\tau$ is
the likelihood threshold and $m_{i}$ is the motion variable associated
to $\mathbf{z}$. 

\subsection{Feature Dimensionality Reduction\label{subsec:Feature-Dimensionality-Reduction}}

For SG no DDR is proposed since the feature space is relatively
small. On the contrary, DDR is beneficial for DGs feature extraction
due to the relatively large feature space and to standardize the variability
of DG length (reducing the gesture length to a small fixed size
\textendash{} resampling). We propose two forms of dimensionality
reduction, using CI and PCA. CI is used to transform any variable-length
DG sample $\mathbf{X}^{(i)}\in\mathbb{M}{}^{d\times n}$ into
a fixed-dimension sample $\mathbf{X}'\in\mathbb{M}^{d\times k}$,
effectively reducing the sample dimensionality if $k<n$. PCA performs
an orthogonal linear transformation of a set of $n$ \emph{$d$}-dimensional
observations, $\mathbf{X}\in\mathbb{M}^{d\times n}$, into a subspace
defined by the principal components (PC). The PC have necessarily a length
smaller than or equal to the number of original dimensions, $d$.
The first PC has the largest variance observed in the data. Each of
the following PC is orthogonal to the preceding component and
has the highest variance possible under this orthogonality constraint.
The PC are the eigenvectors of the covariance matrix and its
eigenvalues are a measure of the variance in each of the PC.
Therefore, PCA can be used for reducing the dimensionality of gesture
data by projecting such data into the PC space and truncating
the lowest-ranked dimensions. These dimensions have the lowest eigenvalues,
so that truncating them retains most of the variance present in the
data, i.e., most of the information of the original data is kept in
the reduced space. In this study, the singular vectors of the samples
across time are calculated and used as features. The first singular
vector determines the direction in the PC-space in which there is
the most significative variance along a DG. This means that the singular
vector is a measure of the relative variance of each variable along
time and good features for DG classification are expected. Another
advantage is that the PC can be calculated even before a DG is finished
(incomplete data), remaining good predictors. As a result, these features
are actually time series since we can calculate them on any window
of data starting with the first frame and ending with any arbitrary
frame of the sample.

\subsection{UC2017 Hand Gesture Dataset}

We introduce the UC2017 static and dynamic gesture dataset. Most researchers
use vision-based systems to acquire hand gesture data. Despite that,
we believe that more reliable results from more complex gestures can
be obtained with wearable sensor systems. There are not many datasets
using wearable systems due to the plethora of data gloves in the market
and their relative high cost. For these reasons, we opted by creating
a new dataset to present and evaluate our gesture recognition framework.
The objectives of the dataset are: (1) provide a superset of hand
gestures for HRI, (2) have user variability, (3) to be representative
of the actual gestures performed in a real-world interaction.

We divide the dataset in two types of gestures: SGs and DGs.
SGs are described by a single timestep of data that represents
a single hand pose and orientation. DGs are variable-length
time series of poses and orientations with particular meanings. Some
of the gestures of the dataset are correlated with specific interactions
in the context of HRI, while the others were arbitrary selected
to enrich the dataset and add complexity to the classification problem.

\begin{figure}
	\includegraphics[width=0.95\columnwidth]{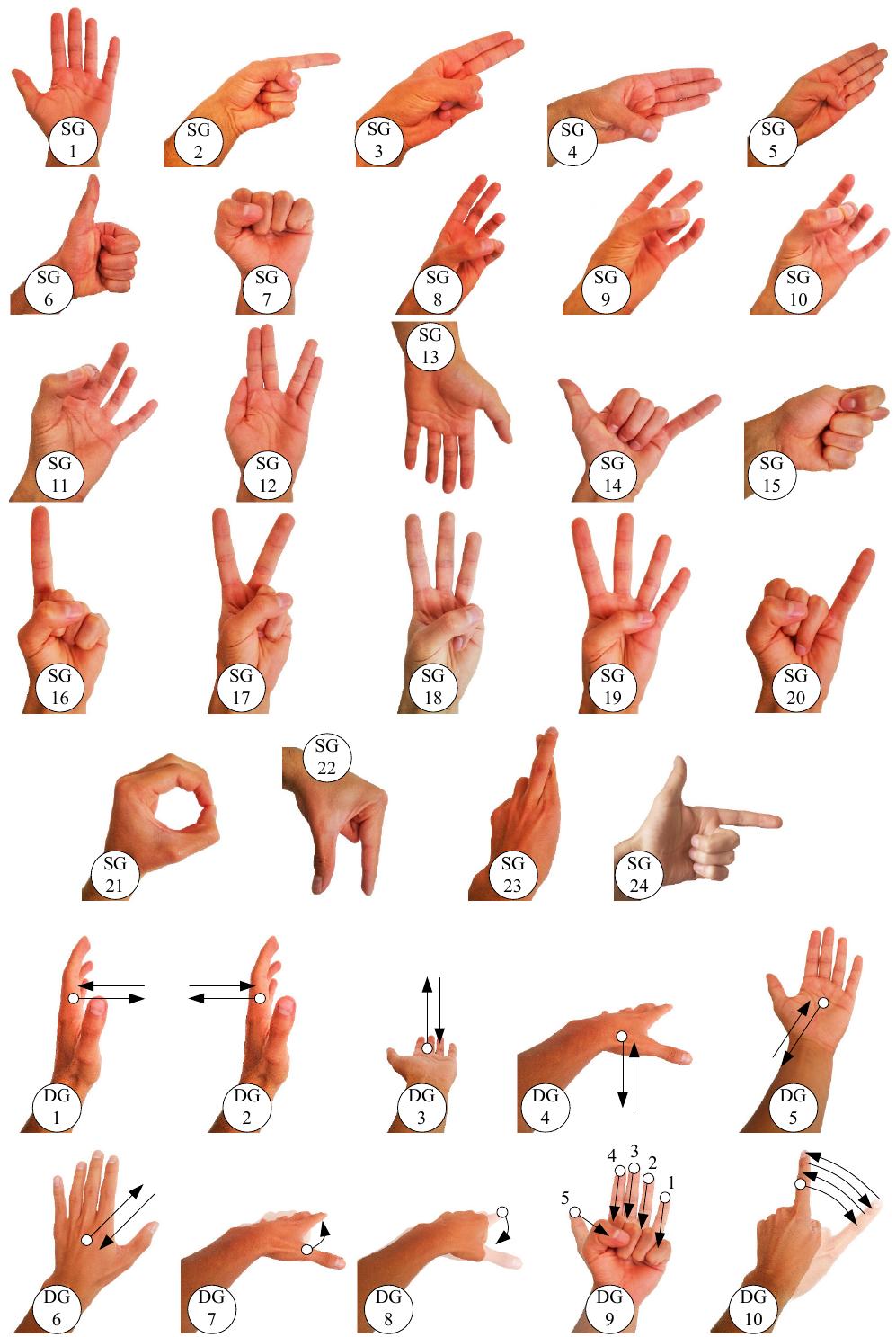}\centering
	
	\caption{Representations of the library of 24 static gestures and 10 dynamic
		gestures of the UC2017 library.\label{fig:ii_gesture_library}}
\end{figure}

The library is composed of 24 SGs and 10 DGs, Fig. \ref{fig:ii_gesture_library}.
The dataset includes SG data from eight subjects with a total
of 100 repetitions for each of the 24 classes (2400 samples in total).
The DG samples were obtained from six subjects and has cumulatively
131 repetitions of each class (1310 samples in total). All of the
subjects are right-handed and performed the gestures with their left
hand.

A data glove (CyberGlove II) and a magnetic tracker (Polhemus Liberty)
are used to capture the hand shape, position and orientation over
time. The glove provides digital signals $g_{i}$ proportional to
the bending angle of each one of the 22 sensors elastically attached
to a subset of the hand's joints: 3 flexion sensors per finger, 4
abduction sensors, a palm-arch sensor, and 2 sensors to measure wrist
flexion and abduction. These 22 sensors provide a good approximation
of the hand's shape. The tracker's sensor is rigidly attached to the
glove on the wrist and measures its position in Cartesian space and
orientation in respect to a ground-fixed frame (a magnetic source
cube defines the reference coordinate system frame). The orientation
is the rotation between the fixed frame and the frame of the tracker
sensor, given as the intrinsic Euler angles yaw, pitch and roll (ZYX).
The Cartesian position $(x,y,z)$ is denominated by $(l_{1},l_{2},l_{3})$
and in terms of orientation the roll angle is denominated by $l_{4}$,
the pitch by $l_{5}$ and the yaw by $l_{6}$. Sensor data are fused
together online since the sensors have slightly different acquisition
rates \textendash{} 100Hz for the glove and 120Hz for the tracker.
Tracker data are under-sampled by gathering only the closest tracker
frame in time.

A goal was to obtain multiple repetitions of each gesture in the library
to build the dataset. We also want the dataset to be representative
of real-world conditions, so we must guarantee that the samples are
independent.

The magnetic tracker reference was fixed in a location free of magnetic
interference. The users are then asked to put on the data glove on
their own on their left hand, even though all of our test subjects
were right-handed. As a result, the sensors are not carefully placed,
which should yield a dataset with larger variance. There is no calibration
done in this setup. The subjects follow a graphical interface that
shows the representation of the gesture to be performed and press
a button to save a sample. The order of the gestures was randomized
to prevent order dependencies. Furthermore, the subjects were requested
to repeat the sampling for two to three different sessions. We have
also implemented an online movement detection algorithm to facilitate
the labeling of DG, namely the starting and ending frames.

A final point should be made about the random sampling of the DG.
We have included in the samples the transition between the ending
pose of the previous sample and the starting pose of a sample (movement
epenthesis). Owing to random sampling, there is a high likelihood
that all of the possible transitions were recorded. The dataset and accompanying code are publicly available at GitHub \url{https://github.com/MiguelSimao/UC2017_Classification}.

Gesture classification must be
independent of the subject's position and orientation in space. Since
the user is free to move around in the world reference frame $\{W\}$,
we need to make sure that every gesture sample has its feature data
reported to their local reference frame $\{L\}$. Origin and orientation
of $\{L\}$ are defined in relation to $\{W\}$ at the instant a gesture
begins. The proposed transformation is composed of a 3D translation
and a rotation around the vertical axis $z$.

\begin{figure}
	\begin{centering}
		\includegraphics[width=1\columnwidth]{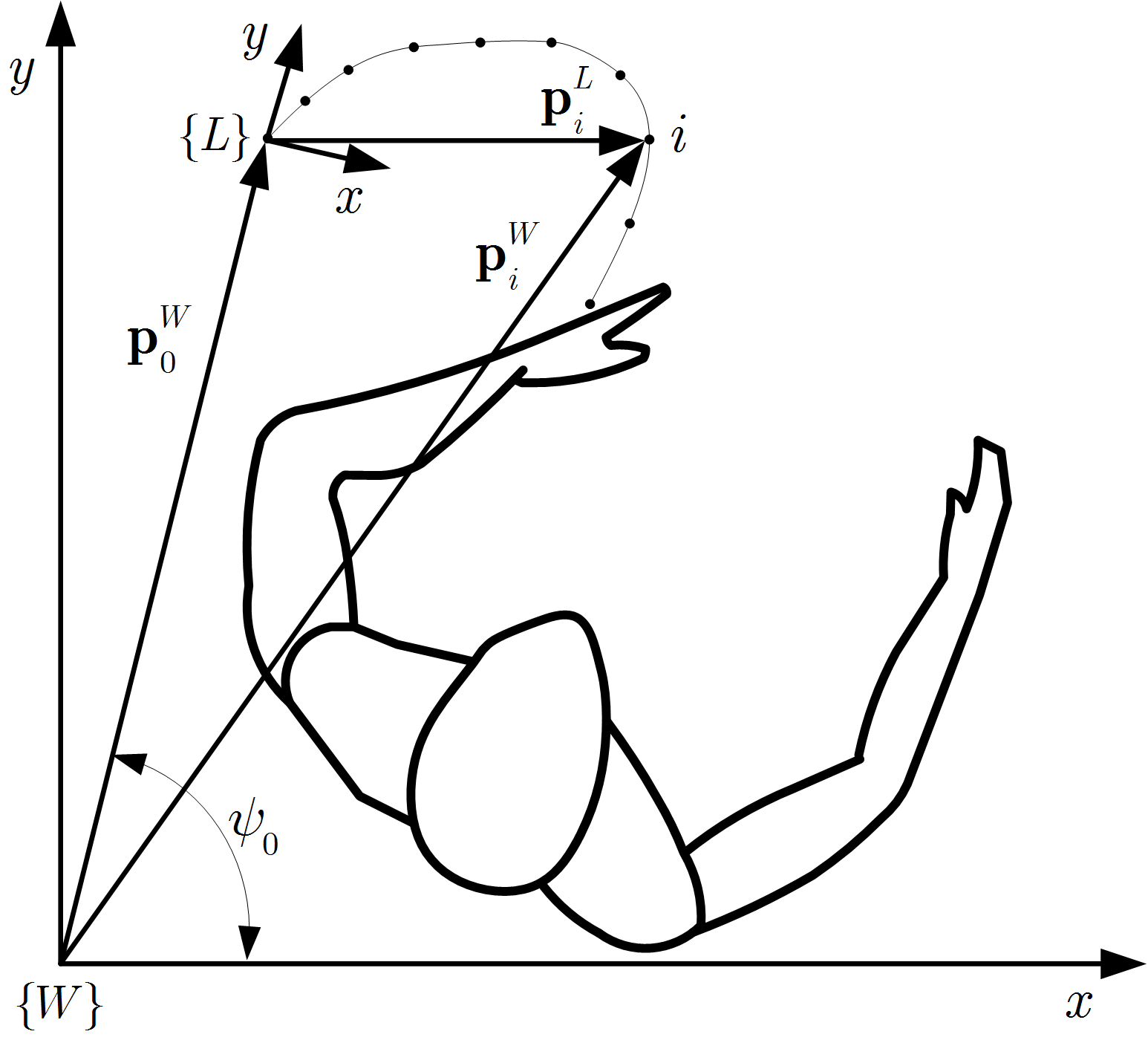}
		\par\end{centering}
	\caption{Representation of the transformation of the world coordinates $\{W\}$
		of a gesture to a local coordinate frame $\{L\}$.\label{fig:transformation}}
\end{figure}

We denote $l_{i}$ and $g_{i}$ as the \emph{i}-th DOF of the
tracker and glove, respectively. At the beginning of a DG sample
$\mathbf{X}^{(i)}$, the yaw $\psi_{0}$ ($l_{6}$) and position $\mathbf{p}_{0}^{W}$
($l_{1}\text{ through }l_{3}$) of the first frame are stored. The
yaw angle is used to calculate the rotation matrix for each sample.
It allows to consistently distinguish directions, such as
right, left, front and back, in respect to the user. The rotation
is applied to every frame of a sample so that we can translate the
coordinate system to frame $\{L\}$, Fig. \ref{fig:transformation}: 

\begin{equation}
	\mathbf{p}_{i}^{L}=\left.\mathbf{R}_{L}^{W}\right.^{T}(\psi_{0})\cdot\left(\mathbf{p}_{i}^{W}-\mathbf{p}_{0}^{W}\right)\label{eq:position}
\end{equation}
where $\mathbf{p}_{i}^{L}$ is the position of the $i\text{th}$ frame
in respect to the local reference frame $\{L\}$, and $\mathbf{R}_{L}^{W}(\psi_{0})$
is the rotation matrix that represents the rotation around $z$ of
the world reference frame $\{W\}$ to $\{L\}$ by $\psi_{0}$ degrees.
The rotation matrix is given by:

\begin{equation}
	\mathbf{R}_{z}(\psi)=\left(\begin{array}{ccc}
		\cos\psi & -\sin\psi & 0\\
		\sin\psi & \cos\psi & 0\\
		0 & 0 & 1
	\end{array}\right)
\end{equation}
After the transformation, the yaw angle is $\psi_{i}^{L}=\psi_{i}^{W}-\psi_{0}^{W}$.
In summary, we have a transformation function $\Psi$ applied to a
DG sample $\mathbf{X}^{(k)}\in\mathbb{M}^{d\times n}$:

\begin{equation}
	\begin{aligned}\Psi:\quad & \R^{6\times\bullet} & \rightarrow & \R^{6\times\bullet}\\
		& \mathbf{X}_{ij}^{(k)} & \rightarrow & \left.\left(\mathbf{p}_{1}^{L}\ \mathbf{p}_{2}^{L}\ \mathbf{p}_{3}^{L}\ \mathbf{l}_{4}\ \mathbf{l}_{5}\ \psi^{L}\right)_{\bullet j}^{(k)}\right.^{T}
	\end{aligned}
	,\quad\begin{array}{c}
		i=23,...,28\\
		j=1,2,...,n
	\end{array}\label{eq:def-transform}
\end{equation}
where $\left(...\right)_{\bullet j}^{(k)}$ corresponds to the \emph{j}-th
timestep of sample \emph{k}.

The dataset is split before feature extraction.
It is shuffled and split in three subsets: training (70\%), validation
(15\%) and test (15\%). The training set was used to train the classifiers
and to obtain feature scaling parameters, such as mean and standard
deviation. The classifiers' hyperparameters were optimized for accuracy
on the validation set. The generalization capability of the model
is measured by the accuracy on the test set. The samples of one subject
were held-out from the training set to ascertain the performance on
new users. Users that trained the system are designated by \textquotedblleft trained
users\textquotedblright{} and users that that did not train the system
are designated by \textquotedblleft untrained users\textquotedblright .

\subsection{Feature Extraction}

For SGs the features are all the angles provided
by the glove, $g_{1}\ g_{2}\ ...\ g_{22}$, and the pitch angle, $l_{5}$
(to differentiate gestures with similar handshapes and distinct orientation).
Thus, the features chosen are simply a subset of the available raw
data. Finally, the features are standardized by $x'_{i}=\left(x_{i}-\bar{x}_{i}\right)/s_{i}$,
where $x'_{i}$ is the standardized value of feature i, $x_{i}$ is
the value of the feature, $\bar{x}_{i}$ and $s_{i}$ are the mean
and standard deviation of the feature in the training set. The validation
and test sets are standardized by these same means and standard deviations. 

For DGs we propose three different sets of features.
For all sets, data samples are preprocessed according to $\Psi$,
defined in (\ref{eq:def-transform}):

\begin{equation}
	\mathbf{X}'{}^{(i)}\text{=}\Psi(\mathbf{X}^{(i)}),\quad\mathbf{X}^{(i)}\in\mathbb{R}^{28\times n}
\end{equation}
where \emph{n} is the length of the DG sample. Starting from $\mathbf{X}'$,
the first proposed set DG-CI uses the full DG data resized to a fixed
length by applying CI. The second set, DG-PV, is based on PCA and
represents the extraction of the first principal vector (PV) from
DG data. The third set is the preprocessed data, which we call RAW.

For DG-CI, given a DG sample $\mathbf{X}^{(i)}$ with $n$ frames
($\mathbf{X}^{(i)}\in\mathbb{R}^{28\times n}$), the goal is to resample
it to a fixed size $n'$. The value for $n'$ can be chosen arbitrarily
but higher values have a detrimental effect on the classification
accuracy. Training the classifier is faster and often better with
less features. For all experiments, the value $n'=20$ was used because
the gesture lengths in the dataset vary between 20 and 224 frames.
The lowest length was selected. Applying CI, the result is a matrix
$\mathbf{Z}\in\R^{28\times20}$. By concatenating every frame vertically,
$\mathbf{Z}$ is transformed into a vector $\mathbf{z}\in\R^{560}$:

\begin{equation}
	\mathbf{z}^{(i)}=\left(\left.\mathbf{Z}_{\bullet1}^{(i)}\right.^{T},\left.\mathbf{Z}_{\bullet2}^{(i)}\right.^{T},\ldots,\left.\mathbf{Z}_{\bullet20}^{(i)}\right.^{T}\right)^{T}\label{eq:def-features-DG1}
\end{equation}

In DG-CI the feature extraction involves the whole DG data, so there
is a prediction only after the gesture is complete. However, it is
beneficial to have an early classification from incomplete data, i.e.,
before the full gesture data are available. For DG-PV, PCA allows
to obtain features from incomplete gesture data and still obtain time-coherent
features. We apply this methodology for each timestep of the gesture.
The feature vector for sample $i$ at timestep $j$ is calculated
by:

\begin{equation}
	\boldsymbol{z}_{j}^{(i)}=\text{\emph{pv}}\left(\left[\mathbf{X'}_{\bullet1}^{(i)}\ \mathbf{X'}_{\bullet2}^{(i)}\ \ldots\ \mathbf{X'}_{\bullet j}^{(i)}\right]\right),\quad j>1
\end{equation}
where \emph{pv} is a function that extracts the principal vector from
its argument. $\mathbf{X'}^{(i)}$ is the standardized sample $\mathbf{X}^{(i)}$,
i.e., with zero mean and unit variance.

A single sample may originate multiple feature vectors depending on
the timestep \emph{$j$}. We used the set $J^{(i)}=\left\{ x:\ 2\leq x\leq n^{(i)}\land x\in\mathbb{N}\right\} $
for training and validation of the classifiers, where $n^{(i)}$ is
the sample length. To simplify the display of results, we tested with
a subset $J^{(i)}=\left\{ \left\lceil 0.25n\right\rceil ,\left\lceil 0.5n\right\rceil ,\left\lceil 0.75n\right\rceil ,\left\lceil 1.0n\right\rceil \right\} $,
where $\left\lceil \right\rceil $ represents the ceiling function.
Simply put, this means that we are testing the features sets extracted
from the samples starting at the first timestep and ending at 25\%,
50\%, 75\% and 100\% of gesture length.

The final step of feature extraction for all features sets is feature
scaling, i.e., the standardization of the features as described for
SGs.

\section{Results and Discussion}

The accuracy of the classifier models on both SG and DG data was obtained
considering a segmentation accuracy estimated to be about 98\%. The
error is mostly oversegmentation, i.e., pauses in the middle of a
DG where the subject slows down or hesitates. In this scenario, the
classification of the DGs is more likely to fail due to lack of gesture
data.

\subsection{Static Gestures}

The accuracy of several classifiers was evaluated on the UC2017 dataset.
The objective is to compare the performance of ANNs with other machine
learning models: K-Nearest Neighbors (KNN), Support Vector Machines
with a Radial Basis Function kernel (RBF SVM), Gaussian Processes
(GP), Random Forests (RF), Gaussian Naive Bayes (NB) and Quadratic
Discriminant Analysis (QDA). The training time, inference time and
accuracy of the models are measured and evaluated as performance parameters.

\begin{table}
	\caption{Training and inference times of several classifiers, accuracy on the
		train, validation and test data subsets for SGs. The test scores are
		divided into the scores of the trained and untrained users (other).\label{tab:results_sg_accuracy} }
	
	\begin{tabular}{r@{\extracolsep{0pt}.}lr@{\extracolsep{0pt}.}lr@{\extracolsep{0pt}.}lr@{\extracolsep{0pt}.}lcr@{\extracolsep{0pt}.}l}
		\toprule 
		\multicolumn{2}{c}{} & \multicolumn{4}{c}{Time (s)} & \multicolumn{5}{c}{Accuracy (\%)}\tabularnewline\cmidrule(lr){3-6} \cmidrule(lr){7-11}
		\multicolumn{2}{c}{Model} & \multicolumn{2}{c}{Train} & \multicolumn{2}{c}{Test} & \multicolumn{2}{c}{Train} & Validation & \multicolumn{2}{c}{Test (other)}\tabularnewline
		\midrule
		\midrule 
		\multicolumn{2}{c}{ANN} & 127&0 & 0&1 & 97&9 & 94.2 & 94&6 (87.9)\tabularnewline
		\multicolumn{2}{c}{QDA} & 0&0 & 0&0 & 99&6 & 91.7 & 94&9 (66.7)\tabularnewline
		\multicolumn{2}{c}{RBF SVM} & 0&1 & 0&2 & 98&0 & 94.2 & 95&2 (83.3)\tabularnewline
		\multicolumn{2}{c}{Gaussian Process} & 23&8 & 13&1 & 99&8 & 991 & 94&9 (69.7)\tabularnewline
		\multicolumn{2}{c}{KNN} & 0&0 & 4&2 & 96&0 & 89.2 & 93&9 (53.0)\tabularnewline
		\multicolumn{2}{c}{Naive Bayes} & 0&0 & 0&0 & 93&4 & 88.3 & 91&2 (69.7)\tabularnewline
		\multicolumn{2}{c}{Random Forest} & 0&1 & 0&0 & 99&8 & 94.7 & 95&6 (92.4)\tabularnewline
		\bottomrule
	\end{tabular}\centering
\end{table}

The ANN used, implemented with Keras, is a feed-forward neural network
(FFNN). Its architecture and hyperparameters were optimized on the
validation dataset using random grid search. Grid search and manual
search are the most common techniques for hyperparameter optimization \cite{Hinton122}. Grid search generates candidates from a grid of
parameter values in which every possible combination of values is
tested to optimize hyperparameters (detailed parameters and Python
code available in supplementary material). The optimal network has
two dense hidden layers of 200 neurons each. Between these layers,
there is a Gaussian noise layer with $\sigma=0.6$. The transfer functions
of the dense layers are linear and rectified. A final layer implements
the \emph{softmax} function to obtain the probability distribution
over the target classes. For weight regularization, we used the L2
distance with a factor of 0.005 and a weight decay coefficient of
$10^{-7}$. The optimization was done using Stochastic Gradient Descent
(SGD) with batches of 32 and a learning rate of 0.001. Furthermore,
in order to prevent overfitting, we used early stopping when there
is a minimum on the validation loss with a tolerance of 10 epochs.
The hyperparameters of the remaining classifiers were optimized using
manual search (detailed parameters and Python code available in supplementary
material).

The performance of the trained classifiers is shown in Table \ref{tab:results_sg_accuracy}.
The best performance on the test set was $95.6\%$ on the trained
users (92.4\% on the untrained), obtained with the RF. The ANN was
slightly worse, with 94.6\% and 87.9\% accuracy on the trained and
untrained users, respectively, leading to the conclusion that in this
case, the RF is generalizing better to new users than the ANN. The
next best performance was the SVM, with 95.2\% accuracy. The other
models performed very well on the trained users, but clearly overfitted
the dataset, since their accuracy on untrained users is below 70\%.

The accuracy of the ANN model for each individual subject varies between
87.9\% (untrained user) and 100.0\%, while one of the trained users
reached only 88.2\%. This subject was one of the authors and was involved
in the definition of the gesture library, which may have originated
samples that are significantly different from those of the other users.
The distribution of errors per class was nearly random, with no gestures
being mixed consistently.

\begin{figure}
	\includegraphics[bb=0bp 0bp 246bp 116bp,width=0.98\columnwidth]{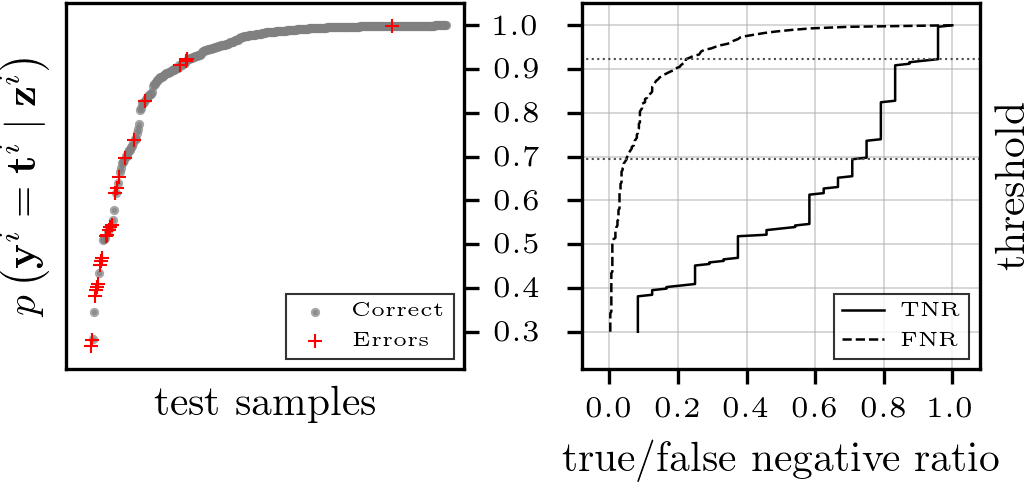}\centering
	
	\caption{On the left, sorted activation values of the winning class for each
		sample of the SG test set. The red crosses correspond to classification
		errors. On the right, the true and false negative ratios (TNR/FNR)
		when we apply a threshold to discard errors. The horizontal dotted
		lines correspond to the 0.696 and 0.923 thresholds.\label{fig:DG-neuron-activation-1}}
\end{figure}

We also present the test results of the feasibility of removing poorly
classified gestures by their score, Fig. \ref{fig:DG-neuron-activation-1}.
This analysis was done with the ANN classifier, since it outputs a
probability distribution over the classes. We present the score of
the winning class $p(\mathbf{y}^{i}=\mathbf{t}^{i}\mid\mathbf{z}^{i}),$
i.e., the probability of the ANN output $\mathbf{y}^{i}$ being the
expected target $\mathbf{t}^{i}$ given the feature vector $\mathbf{z}^{i}$.
It is impossible to define a score threshold to exclude misclassifications
without excluding also some good ones. The trade-off is demonstrated
in Fig. \ref{fig:DG-neuron-activation-1} on the right, via the true
and false negative ratio. If we agree that a $5\%$ False Negative
Ratio (FNR) is acceptable, the threshold $0.696$ reduces miss-classifications
by $71\%$. On the other hand, if we want $95\%$ of miss-classifications
to be discarded, we lose $22\%$ of valid classifications with a score
threshold of $0.923$. The latter is not a particularly good trade-off.
In this context, another solution should be found, such as generating
false samples so that the classifier learns how to better separate
them.

\subsection{Dynamic Gestures}

From the UC2017 dataset, four sets of data contemplating different
features were considered for the experiments, Table \ref{tab:results_sg_accuracy-1}.
CI-FULL and PV-FULL output a single classification per DG, while PV-TS,
RAW-LSTM and RAW-CNN output a classification for each timestep of
a DG.

\begin{table}
	\caption{Feature sets considered for the experiments. CI refers to cubic interpolation,
		PV to principal vectors and RAW to no feature extraction.\label{tab:results_sg_accuracy-1} }
	
	\begin{tabular}{r@{\extracolsep{0pt}.}l>{\centering}p{0.72\columnwidth}}
		\toprule 
		\multicolumn{2}{c}{Feature set} & Description\tabularnewline
		\midrule
		\midrule 
		\multicolumn{2}{c}{CI-FULL} & CI applied to raw preprocessed data describing a full DG sample\tabularnewline
		\multicolumn{2}{c}{PV-FULL} & PVs applied to raw preprocessed data describing a full DG sample\tabularnewline
		\multicolumn{2}{c}{PV-TS} & PVs applied sequentially to a DG sample, starting from its first frame
		to an arbitrary timestep \tabularnewline
		\multicolumn{2}{c}{RAW-LSTM} & Raw preprocessed data classified by LSTMs\tabularnewline
		\multicolumn{2}{c}{RAW-CNN} & Raw preprocessed data classified by CNNs\tabularnewline
		\bottomrule
	\end{tabular}\centering
\end{table}

\begin{table}
	\caption{Classification accuracy for the full DG experiments. The test scores
		are divided into the scores of the trained and untrained (other) users.
		\label{tab:DG-FULL-ACCURACY}}
	
	\begin{tabular}{cccccccc}
		\toprule 
		& \multicolumn{3}{c}{CI-FULL} &  & \multicolumn{3}{c}{PV-FULL}\tabularnewline
		\cmidrule{2-4} \cmidrule{6-8} 
		Model & Train & Val & Test (other) &  & Train & Val & Test (other)\tabularnewline
		\midrule
		\midrule 
		ANN & 100.0 & 98.5 & 99.3 (96.2) &  & 99.7 & 91.3 & 94.4 (66.0)\tabularnewline
		LDA & 100.0 & 98.0 & 97.2 (92.5) &  & 67.2 & 60.7 & 68.8 (50.9)\tabularnewline
		KNN & 97.9 & 94.4 & 96.5 (86.8) &  & 90.0 & 81.1 & 84.7 (62.3)\tabularnewline
		RF & 100.0 & 98.0 & 97.2 (86.8) &  & 100.0 & 92.3 & 88.9 (67.9)\tabularnewline
		SVM & 99.8 & 98.5 & 97.9 (96.2) &  & 87.9 & 82.1 & 79.2 (60.4)\tabularnewline
		\bottomrule
	\end{tabular}\centering
\end{table}

The results for the experiments using CI-FULL features are
shown in Table \ref{tab:DG-FULL-ACCURACY}. Multiple classifiers were
tested and the results report the accuracy achieved by the best ones.
The hyperparameters were chosen by manual search for all the classifiers
(detailed parameters and Python code available in supplementary material).
As an example, the ANN has two hidden layers of 100 and 200 nodes
each, their activation function is linear and rectified, and the output
is the \emph{softmax} function. The weights were regularized using
the L2 distance. For optimization, SGD was used with a batch size
of 128 and a learning rate of 0.01.

The accuracy of the classifiers is generally excellent, around 97.0\%,
in the test set for trained users and up to 96.2\% for untrained users.
The KNN and RF classifiers did not generalize as well to new users,
reaching an accuracy of just 86.8\%. This is most likely explained
by the size of the feature vector (560) and comparatively low number
of samples. On the other hand, the SVM performed nearly as well as
the ANN on untrained users (96.2\%), but worse on trained users (99.3\%
vs 97.9\%).

\begin{figure}
	\includegraphics[width=1\columnwidth]{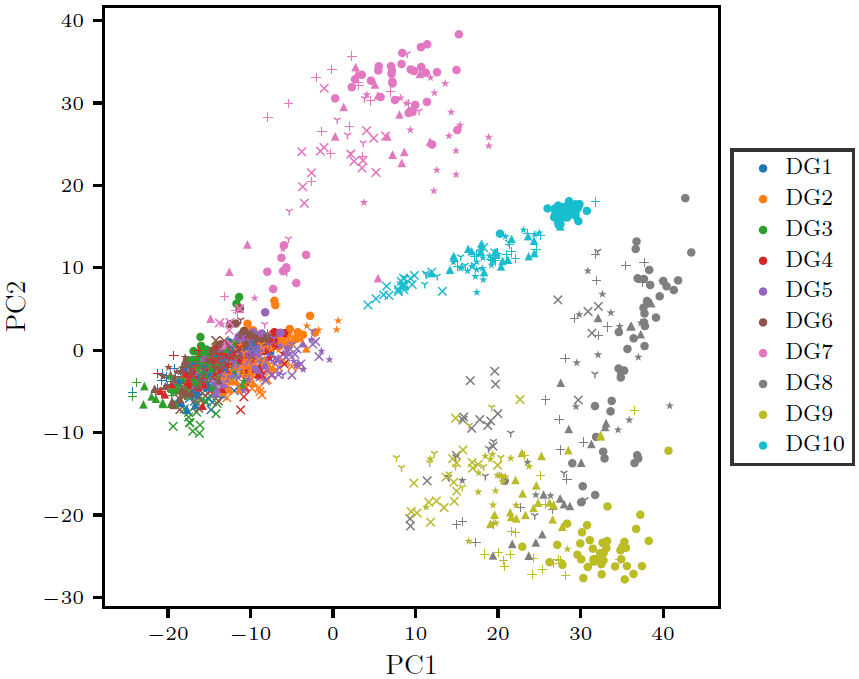}\centering
	
	\caption{DGs projected on the plane defined by the first two principal components
		of the entire dataset - training, validation and test data). The ten
		DG classes are represented by different colours and markers. From
		DG1 to DG6 we can observe a cluster and then for DG7, DG8, DG9 and
		DG10 we have other clusters. This is because the gestures from DG1
		to DG6 are all performed with the hand open (no variations in finger
		angle data). \label{fig:INTRO-Segmentation-1-1}}
\end{figure}

\begin{table*}
	\caption{DG classification sequential accuracy for the time-series based experiments
		PV-TS, RAW-CNN and RAW-LSTM at 25, 50, 75 and 100\% of DG completion.
		The test scores are divided into the scores of the trained and untrained
		(other) users. \label{tab:DG-TS-ACCURACY}}
	
	\begin{tabular*}{1\textwidth}{@{\extracolsep{\fill}}cr@{\extracolsep{0pt}.}lr@{\extracolsep{0pt}.}lcccccccccccccc}
		\toprule 
		\multirow{2}{*}{Model} & \multicolumn{4}{c}{Time (s)} & \multicolumn{4}{c}{Train accuracy (\%) } &  & \multicolumn{4}{c}{Validation accuracy (\%) } &  & \multicolumn{4}{c}{Test accuracy (\%) }\tabularnewline
		\cmidrule{6-9} \cmidrule{11-14} \cmidrule{16-19} 
		& \multicolumn{2}{c}{Train} & \multicolumn{2}{c}{Test} & 0.25 & 0.50 & 0.75 & 1.00 &  & 0.25 & 0.50 & 0.75 & 1.00 &  & 0.25 & 0.50 & 0.75 & 1.00\tabularnewline
		\midrule
		\midrule 
		ANN & 94&7 & 0&2 & 95.3 & 96.9 & 95.4 & 91.1 &  & 74.5 & 86.7 & 88.3 & 85.2 &  & 77.8 (49.1) & 91.0 (75.5) & 88.9 (71.7) & 84.7 (62.3)\tabularnewline
		KNN & 25&6 & 3&4 & 99.9 & 99.9 & 100.0 & 99.5 &  & 67.9 & 82.1 & 83.2 & 78.6 &  & 70.8 (43.4) & 83.3 (69.8) & 81.2 (69.8) & 81.9 (54.7)\tabularnewline
		RF & 54&6 & 0&2 & 100 & 100.0 & 100.0 & 100.0 &  & 74.5 & 88.8 & 91.3 & 86.7 &  & 70.1 (54.7) & 88.9 (75.5) & 87.5 (67.9) & 80.6 (67.9)\tabularnewline
		SVM & 102&4 & 8&8 & 97.8 & 97.8 & 97.1 & 93.1 &  & 73.0 & 83.7 & 87.2 & 83.2 &  & 75.7 (41.5) & 89.6 (71.7) & 88.9 (69.8) & 81.2 (60.4)\tabularnewline
		\midrule
		LSTM  & 390&5 & 69&5 & 87.9 & 96.8 & 99.8 & 99.8 &  & 78.1 & 92.9 & 97.4 & 98.5 &  & 81.7 (50.9) & 95.1 (74.5) & 97.2 (87.3) & 96.5 (89.1)\tabularnewline
		CNN & 66&4 & 2&7 & 86.6 & 96.5 & 97.1 & 88.7 &  & 81.1 & 92.9 & 97.1 & 88.7 &  & 84.7 (60.4) & 94.4 (84.9) & 92.4 (73.6) & 81.9 (54.7)\tabularnewline
		\bottomrule
	\end{tabular*}\centering
\end{table*}

\begin{figure*}
	\begin{centering}
		\includegraphics[width=0.95\textwidth]{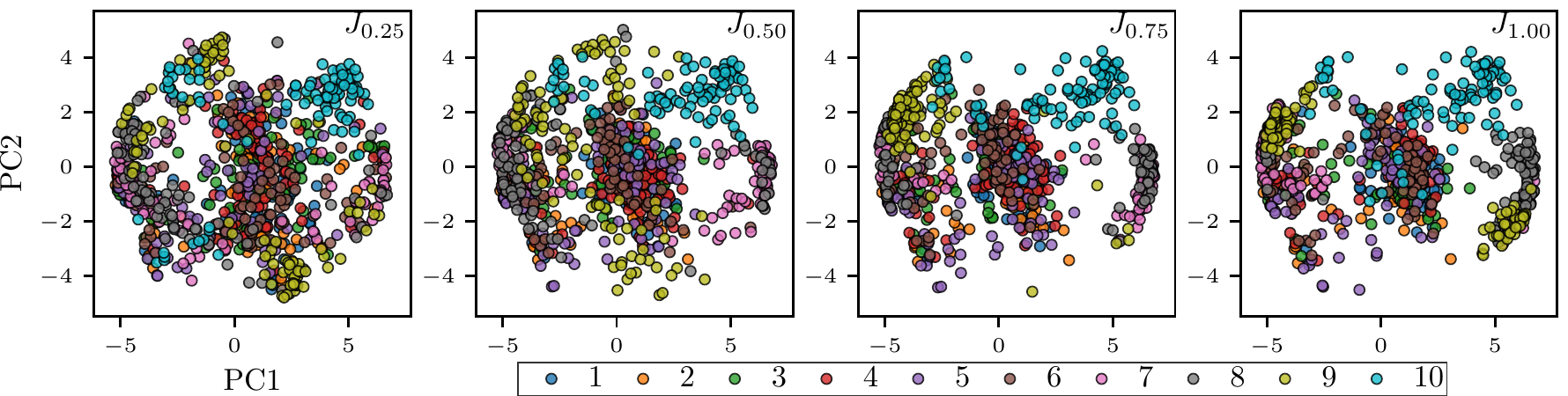}
		\par\end{centering}
	\caption{Plots of the features of the DG training set, with 25\%, 50\%, 75\%
		and 100\% of the data, in a reduced 2D principal component space.
		Each color represents a different class. Results for each classifier
		are detailed in Table \ref{tab:DG-TS-ACCURACY}. \label{fig:DG2-PC-Features}}
\end{figure*}

The results for the experiments using PV-FULL features are
shown in Table \ref{tab:DG-FULL-ACCURACY}. Fig. \ref{fig:INTRO-Segmentation-1-1}
shows the ten DGs projected on the plane defined by the first two
principal components of the data. We have tested the same classifiers
as in CI-FULL, so that we can establish a comparison between the two
feature sets. The accuracy is generally and markedly below of those
obtained by CI-FULL. The ANN accuracy decreased by 4.9\% for trained
users, even with an updated architecture composed by two hidden layers
with 300 units each. The generalization to untrained users was poor,
with just 66.0\% accuracy. The RF achieved slightly better results
on the training and validation sets than the ANN, but the test accuracy
was lower (88.9\%). All of the other models performed significantly
worse. As a conclusion, the PV features lose more information about
the DGs than the CI features, making classification harder. However,
the PV features can be calculated with an arbitrary number of frames
of data, without the full gesture data, therefore allowing sequential
(online) classification. 

The results for the experiments using PV-TS features are
shown in Table \ref{tab:DG-TS-ACCURACY}. For this case we present
the results for the best performing models, a feed-forward ANN, KNN,
RF and SVM. The classifiers' hyperparameters were optimized again
by manual search, but they remained nearly the same as for PV-FULL,
since the problem is similar. However, the ANN architecture was updated
to two hidden layers of 512 and 256 nodes. There is also a $0.1\sigma$
Gaussian noise layer after each hidden layer and a dropout layer (50\%
rate) in between. These noise layers help the network generalize better
at all timesteps, since there is now a feature vector for each timestep
of each DG sample, which in turn originates much more variability
in the features.

We present the training time for each model and the total inference
time for the whole dataset. The accuracy results for each dataset
split are presented and the test split was divided into trained and
untrained users. According to our designation for incomplete data,
the columns ``$0.25$'' correspond to the PV feature vectors calculated
with the first 25\% of timesteps of each DG sample. The same logic
applies to ``0.50'' , ``0.75'' and ``1.00''. The column ``$1.00$''
corresponds to all of the DG timesteps being used, which is the same
case as PV-FULL. However, in PV-TS we use precisely the same classification
model for all timesteps.

The best classification accuracy on the test set was obtained with
the ANN at all timesteps of the test set. The RF and SVM models reached
nearly the same accuracy as the ANN, with 90\% at the middle of the
DGs and 89\% accuracy at 75\% of DG completion. At the end of the
DGs (100\% completion), the accuracy is lower for these models than
for the ANN (81\% vs 85\%). The KNN model performed worse at most
timesteps, reaching only 83.3\%, 81.2\% and 81.9\% accuracy at 50\%,
75\% and 100\% of DG completion, respectively. Interestingly, for
all of the classifiers, the accuracy is better at 50\% and 75\% of
DG completion than at 100\%. This is likely due to the way most of
the gestures of the library have a motion in one direction and then
move back to the initial position. This would mean that the first
half of the gesture is a better predictor of the whole gesture than
the second half. For example, the second half of DG1 is very similar
to the first half of DG2. To aid in visualization, the features are
shown in a 2D PC space (for the test set) in Fig. \ref{fig:DG2-PC-Features}.
There is considerable noise at $J_{0.25}$, which is unfavorable for
classification. However, after that, we see stable clusters, such
as classes 8 and 10. On the other hand, class 9 has many samples flipping
their position at $J_{1.00}$, which may help explain the drop in
accuracy.

\begin{figure*}
	\includegraphics[width=0.74\textheight]{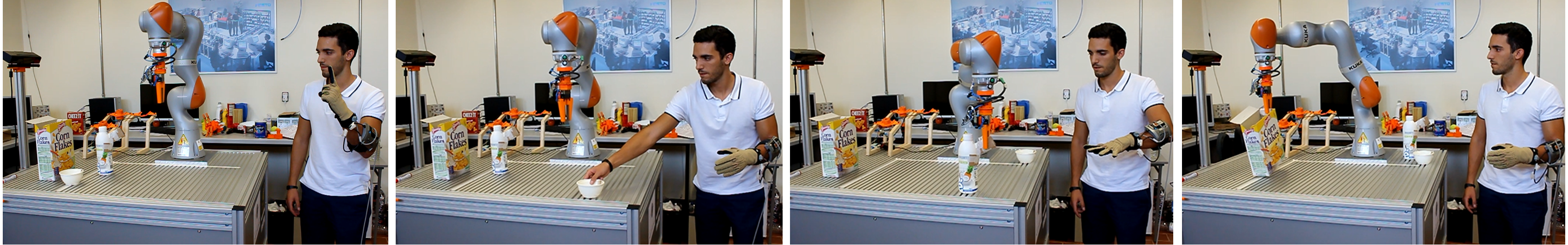}
	
	\caption{The human collaborates with the robot to prepare the breakfast meal.
		The video is available in supplementary material.\label{fig:DEMO-Process}}
\end{figure*}

RAW-LSTM experiments use a LSTM recurrent neural network
that is composed of cells that have memory which can be kept or forgotten
over-time, depending on the sequence of input data. The LSTM structure
and hyperparameters were obtained by manual search. There is a densely
connected layer with 512 units and a $0.4\sigma$ Gaussian noise layer,
which is followed by a LSTM layer of 256 cells. After that we implement
a 50\% dropout layer. The output layer has a \emph{softmax} activation
function. All the other layers have the hyperbolic tangent as transfer
function. Additionally, we increase the weight of the timesteps after
50\% of sample completion, so that the model optimizes accuracy at
later stages of the gesture. Detailed parameters and Python code is
available in supplementary material. In terms of accuracy on the trained
users of the test split, the LSTM outperforms all the other models
at 50\%, 75\% and 100\% of gesture completion, with 95.1\%, 97.2\%
and 96.5\% accuracy, respectively. In respect to generalization to
new users, the LSTM is significantly better than all other models
when 75\% or more data is available. At 75\% and 100\%, the accuracy
on new users is 87.3\% and 89.1\%, respectively. It compares favorably
to the second best performer (ANN) at the cost of a significant increase
in training and inference times, due to the large number of parameters
of the LSTM model. 

We performed the RAW-CNN experiment in the same conditions
as the previous RAW-LSTM. The CNN used had an initial dense hidden
layer with 512 nodes and a \emph{tanh }transfer function. Afterwards,
there are two convolutional (1D) layers with 100 filters each, windows
of 5 timesteps and rectified linear units as transfer function. Each
convolutional layer is followed by Gaussian noise layers of strength
$0.2\sigma$. The model is trained by SGD with a learning rate of
0.001. While oftentimes a CNN can be used to model sequential data
with performance similar to that of a LSTM, in this case it was worse
in accuracy at the later stages of the DGs, and also in terms of generalization
to new users. Considering 50\% of gesture data, the accuracy on trained
users (94.4\%) is close to the LSTM's 95.1\%. The CNN also generalizes
better at that timestep than the LSTM (84.9\% vs 74.5\%). However,
for 75\% of gesture completion, it is significantly worse than the
LSTM for all users, and at 100\% it is worse than the feed-forward
ANN with just 81.9\% and 54.7\% accuracy for trained and untrained
users, respectively. A decrease of this magnitude was unexpected and
it is possibly due to the data padding that occurs during the convolution
operations. The last timestep of a DG is evaluated by a convolution
operation on a window centered on that timestep. However, since the
window has length 5, that means that 2 timesteps correspond to CNN
padding, which is generated, unreliable data, leading to the low score.

The LSTM and CNN approaches have the advantage that the raw sensor
data can be fed directly into the model after scaling, unlike all
of the others, which require carefully chosen feature extraction methods.
For LSTMs, the disadvantage is that the training and inference time
are one to two orders of magnitude higher than the other classifiers,
due to model complexity. The latter is more concerning because classification
must be done online for an efficient human-robot interactive process.
The experimental setup demonstrated that the inference time per frame
for the LSTM model is about 0.16 ms (about 6300 Hz) on a GPU, so it
could be an issue for its implementation in embedded systems.

\subsection{Robot Interface}

We have implemented a gesture-based human-robot interface composed
by gestures from the UC2017 dataset. Since there are delays in data
acquisition, data stream segmentation, candidate sample preprocessing,
classification, decision making and robot communication, we estimate
that the total delay between the end of a gesture performed by the
human and the robot reaction is about 300 ms.

The collaborative robot is a 7 DOF KUKA iiwa equipped with the Sunrise
controller and interfaced using the KUKA Sunrise Toolbox for MATLAB
\cite{safeeaKST}. The attempted collaborative robotic task consists
in preparing a breakfast meal composed by subtasks such as grasping
a cereal box, a yogurt bottle, and pouring the contents into a bowl,
Fig. \ref{fig:DEMO-Process}. These tasks were performed by direct
robot teleoperation, being the robot actions controlled online by
the human gestures and the collaborative process managed by a collaborative
robot task manager \cite{indin2018}. The task manager can be setup
with a number of required validations so that when a gesture is wrongly
classified the system actuates to avoid any potential danger for the
human and/or the equipment. From the library of 24 SGs and 10 DGs,
three subjects were taught the mapping between gestures and specific
robot commands, such as: stop motion, move along X, Y or Z in Cartesian
space, rotate the robot end-effector in turn of X, Y or Z, open/close
the gripper, and teleoperate the robot in joystick mode. Anytime the
user is not performing a given gesture the system is paused.

All users indicated that the interaction process is very natural,
since they can easily select the desired operation modes and the system
is intuitive to use. During the interactive process, the reached target
points can be saved and used in future robot operations. The impedance
controlled robot compensates positioning inaccuracies, i.e., the users
can physically interact with the robot to adjust positioning. Concerning
safety, the subjects indicated that they feel safe in interacting
with this robot due to the fact that the KUKA iiwa is a sensitive
robot that is able to stop its motion when a pre-defined contact force
is reached.

\section{Conclusion and Future Work}

This paper presented an online static and dynamic gesture recognition
framework for HRI. Experimental results using the UC2017 dataset showed
a relatively high classification accuracy on SGs without feature extraction.
For DGs, the use of CI features resulted in high offline classification
accuracy using a regular ANN model. The achieved accuracy compares
favourably with standard classifiers and with deep learning LSTM on
online classification with PCA features. The LSTM uses scaled raw
data, therefore being more easily extendable to new datasets. Nevertheless,
The LSTM degrades when we compare the training and inference time,
which is critical for online implementation. The sequential classification
of DGs with either PCA features or raw data showed an accuracy that
is higher with partial gesture data (50\% or 75\% of the initial frames
of a DG sample) than with the full DG data. In this context, DGs can
be accurately classified in anticipation, even before the user finishes
the gesture in real world, thus allowing faster and more efficient
gesture-based control of a robot by cutting the processing time overheads.

The human-robot interactive process demonstrated that is feasible
to associate the recognized gesture patterns to robot commands and
by this way teleoperate the collaborative robot in an intuitive fashion. 

Future work will be dedicated to testing the proposed solution with
other interaction technologies (vision, IMUs, and EMG). The promising
results obtained with the classification from incomplete data will
be explored as a way to anticipate robot reaction to human commands.

% References

\bibliographystyle{Bibliography/IEEEtranTIE}
%\bibliography{Bibliography/IEEEabrv,Bibliography/BIB_1x-TIE-2xxx}\ %IEEEabrv instead of IEEEfull

\end{document}